\documentclass{article}
\usepackage{acra}
\usepackage[flushleft]{threeparttable}
\usepackage{graphicx}
\usepackage[font=small]{caption}
\usepackage{amsmath,lipsum}
\newcommand{\mypm}{\smash{%
\raisebox{0.35ex}{%
            $\underset{\raisebox{0.5ex}{$\smash -$}}{\smash+}$%
            }%
        }%
}

\title{
Design of a sensing module for a kiwifruit flower pollinator robot
}

\author{Mahla Nejati$^{1}$$^{*}$, Ho Seok Ahn$^{1}$, Bruce MacDonald$^{1}$ \\ $^{1}$The Centre for Automation and Robotic Engineering Science, \\the University of Auckland, Auckland, New Zealand \\ 
$^{*}$mnej691@aucklanduni.ac.nz}

\begin{document}
    
    \maketitle
    \begin{abstract}
     This paper describes steps taken to develop a sensing module for a robotic kiwifruit flower pollinator, which could be integrated into an imaging module and a spray module. The paper described different indicators to present the performance of the sensing module that can be used as a benchmark. The sensing module provides data for the imaging module to detect kiwifruit flower reliably and accurately in the canopy. Four major challenges for a sensing module is the speed, accuracy, visibility, and robustness to variable lighting conditions. Regarding these issues, Basler acA1920-40uc camera with an LM6HC lens were selected from a list of fast cameras and lenses based on different parameters. 
     The sensing module was tested in four orchards and captured 9128 images. According to the saturation rate parameter, the captured images were typical in 96\% of conditions and the rest were glare due to the sunny weather and early season. The camera speed and field of view guarantee that in the highest speed of the robot a flower can be seen at least in three images in normal conditions. The sensing module was calibrated with less than 3mm error and integrated to the spray module. The pollinator module was mounted on a robot and tested in the real world and achieved 79.5\% hit rate at an average velocity of 3.5 km/h.
    \end{abstract}
    \section{Introduction}
    Kiwifruit has a top place in horticultural exports in New Zealand. Export earnings from kiwifruit in 2017/2018 were 1.7 billion dollars \cite{Zesperi2017-18}. The goal is growing global kiwifruit sales revenue to \$4.5 billion by 2050. Achieving the goal requires growing good quality kiwifruit with an efficient method.
    
\begin{figure}[!htbp]
        \centering\includegraphics[width=8.2cm, height=6.5cm]{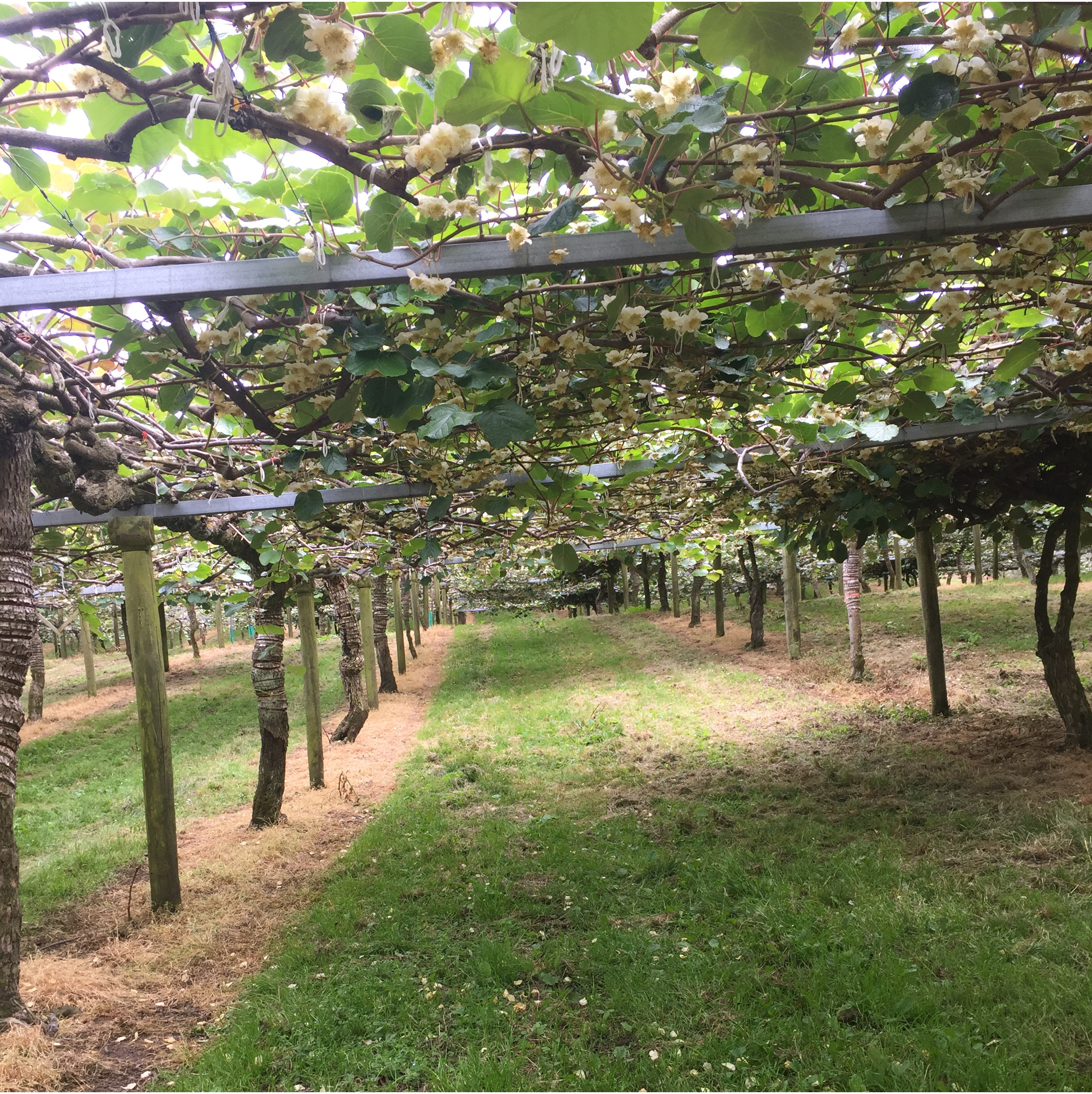}
        \caption{A kiwifruit flower orchard}
        \label{label1}
    \end{figure}

    Male and female kiwifruit plants are required to set a kiwifruit. Alongside plants with kiwifruit flowers, bees should transfer the pollen from the male flower to the female one. Kiwifruit flowers do not have nectar to attract bees. Thereby, orchardists are using artificial methods such as pollen blowers, dusters, and spray dispensers. These methods are costly and need a large human labour force. The kiwifruit flower orchard is shown in figure \ref{label1}. An autonomous kiwifruit pollinator that can spray individual flowers reliably and sustainably in the kiwifruit orchard is beneficial for the New Zealand kiwifruit industry.
    
    The robot consists of a pollinator platform which is mounted on an Autonomous Multi-purpose Mobile Platform (AMMP) \cite{Barnett2017,Jones2019}. AMMP is the developed version of the work discussed in \cite{Scarfe2012a}. In other studies, AMMP was used with a harvesting platform to pick kiwifruit \cite{Williams2019H1,Williams2019H2}.
    
\begin{figure}[!htbp]
        \centering\includegraphics[width=82mm, height=30mm]{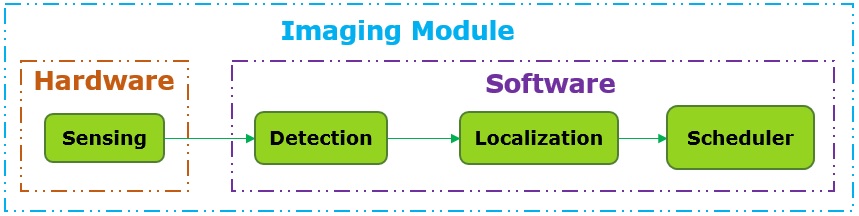}
        \caption{Imaging module diagram}
        \label{fig:pollinatorDiagram}
    \end{figure}

    There is no published research on developing a kiwifruit flower pollinator, so the proposed robot is the first prototype in the world. The pollinator platform contains an imaging module and a spray module. The imaging module consists of hardware and software parts and it is illustrated in figure \ref{fig:pollinatorDiagram}. The hardware part includes a sensing module which captures images. The images are sent to the software part which includes a detection module, a flower localization module, and a flower scheduler module. The detection module finds the kiwifruit flowers and the 3D position of them is calculated by flower localization module and sends to a flower scheduler module. The flower scheduler module sorts and updates flower positions and the scheduler sends their corresponding nozzle number to the spray module, and it fires toward the flower centre. An overview of all modules was presented in \cite{Williams2019} and the aim of this paper is describing and evaluating the sensing module with more details. 
    
    A set of limitations is defined based on the nature of the kiwifruit orchard, the robot design, and the pollination process. The sensing module should produce good quality images and accurate 3D positions. One of the main challenges of the sensing module of a kiwifruit flower pollinator is working in the outdoor environment \cite{Yang2019}. This paper will present the design of the sensing module that can fulfil all requirements. Then, some indicators are defined to describe the performance of the sensing module.
    
    The remainder of the paper is organised as follows: section \ref{sec:Related work} describes the related work for selecting sensors and sensing module design. In section \ref{sec:SystemDesign}, the design of the sensing module is presented by describing the sensor selection, sensor placement. The robot performance is discussed in section \ref{sec:Experiments}. A conclusion is drawn in section \ref{sec:conclusion}.
    \section{Related Work}\label{sec:Related work}
    Generally, one of the critical elements of the vision and localisation system is selecting the suitable sensor by which objects can be detected and localised reliably. Various sensors were used for fruit recognition in the orchard such as colour cameras, multi and hyperspectral cameras, thermal cameras, and a fusion of sensors.
    
    There are some pros and cons of using these sensors. Colour cameras provide colour, geometric and texture information which is typical for identifying the fruit \cite{Dias2018,Bargoti2016,Zemmour2017}.
    However, the various ambient lighting condition is one of the main challenges that have a high impact on the quality of images and was reported as a significant challenge in using colour cameras \cite{Bargoti2016}.
    
    Spectral cameras work based on the absorption and reflection of radiation in specific bands of the spectrum and provide colour and non-visible spectral information. The uneven lighting conditions cause variations in reflected light from objects, and this was reported as a problem when using spectral cameras \cite{Hiroma2002}. However, these cameras have the potential to identify the fruit correctly with a similar colour to the background. If the image capture time and the cost are not important, a spectral camera may be a good option that supplies spectral and colour information.
    
    Colour and spectral cameras are not able to work reliably under rough illumination conditions. To overcome the dynamic lighting condition, a thermal camera can be used for image acquisition. The camera senses the temperature difference between the fruit and the background, and the temperature is attributed to the physical structure and characteristics. This camera can be useful during daytime since fruit absorbs more heat than leaves and trunks. Also, the temperature measurements are not characterised by the fruit colour \cite{Bulanon2011}.
    
    The second part of the vision system is finding the 3D location of the detected fruit. There are a range of sensors that provide the 3D position such as time of flight(ToF), laser ranging finder and structured light 3D sensor \cite{Gongal2015,Narvaez2017}. Besides these sensors, there are some other sensors which need a technique for estimating the 3D position such as a stereo pair. Among all sensors, the laser ranging sensor has the lowest error (3 mm), but it is slow and needs mechanical movement, so it is not suitable for a real-time application \cite{Narvaez2017}. ToF cameras \cite{Silwal2017} and structured light cameras \cite{Nguyen2016} have a reasonable error, 6 mm and 3 mm respectively. However, they have a high sensitivity to variable lighting conditions and have a low frame rate. Hence, the performance depends on the foliage coverage and the position of the sun related to the sensor and the robot velocity. The stereo camera pair can satisfy the real-time requirements \cite{Si2015}. However, it has some disadvantages which are a) the variation in lighting conditions causes poor quality images (overexposed images) b) accurate camera calibration is needed. 
    \section{System Design}\label{sec:SystemDesign}
    The goal is designing a sensing module such that the output of the module can be used for kiwifruit flower detection \cite{Lim2019} and localisation. The sensing module should be integrated with the imaging and spray module. Two major parts of the design are selecting a suitable sensor and determining the geometry of the sensing module. Then, the performance of the sensing module in challenging conditions will be discussed. 
    
\begin{figure}[!htbp]
        \centering\includegraphics[width=3.2in, height=6cm]{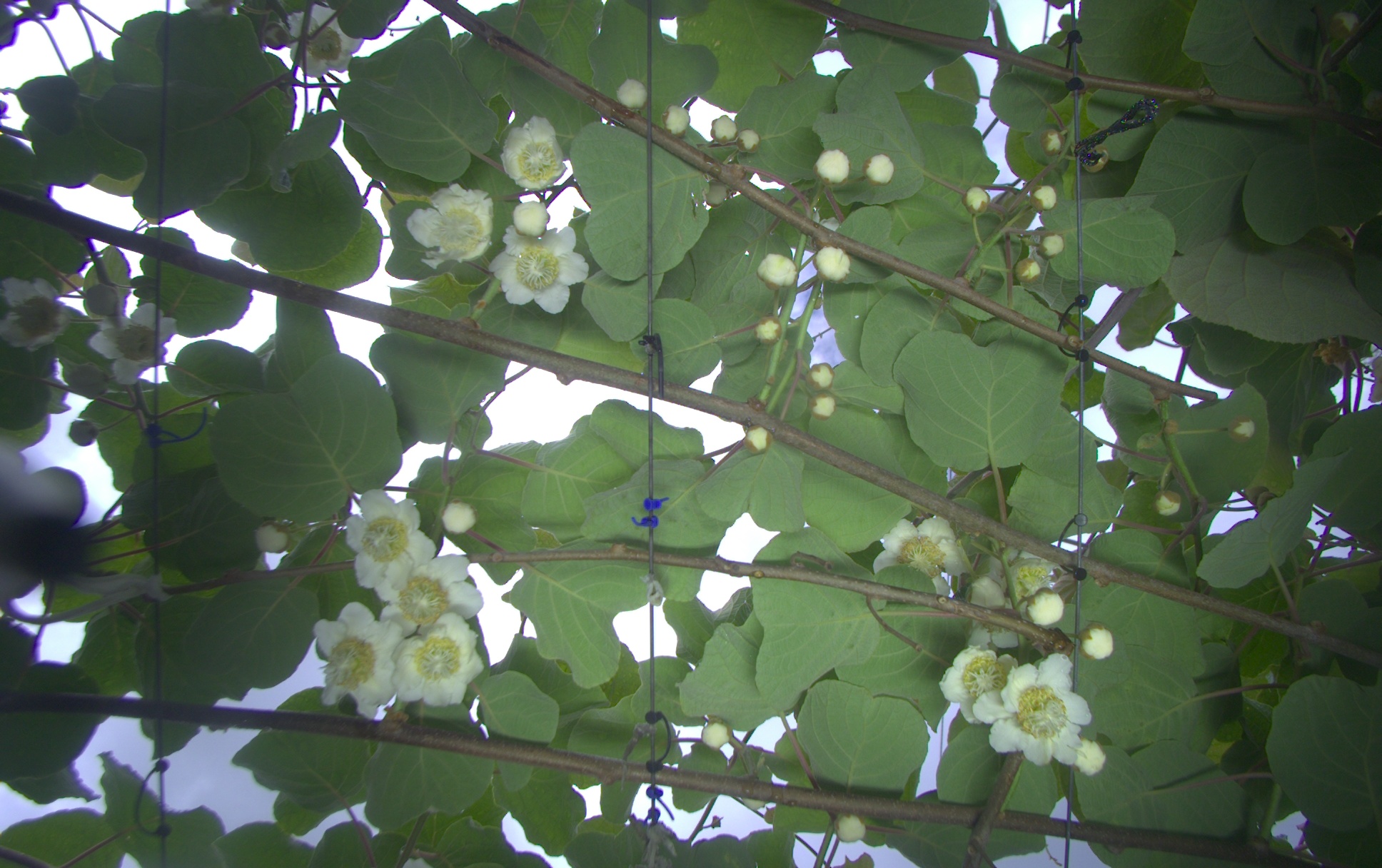}
        \caption{ An example of captured image from the kiwifruit flower orchard using the color camera}
        \label{label2}
    \end{figure}

    With the aim of choosing a suitable sensor, various parameters should be considered which are categorised by the design of the robot and the orchard environment. The sensor should be fast and able to work reliably under variable lighting conditions. The kiwifruit orchard is built on a pergola structure which makes kiwifruit flowers hang downward in the canopy (figure \ref{label1}).
    The robot travels underneath the canopy, so cameras face the sun. It can be concluded that only stereo cameras can meet the limitations due to the conditions discussed in section \ref{sec:Related work}. An example of the captured image using the colour camera from a kiwifruit flower orchard is shown in figure \ref{label2}.
    \subsection{Camera Selection}
    The design of the robot imposes some restrictions on the parameters that are vital for camera selection. These parameters are camera placement space, working distance, and the maximum distance from the cameras to nozzles. Alongside these parameters, the number of pixels that represents flower at the working distance is essential. The performance of current detection methods is related to the size of the object. More features can be extracted from a larger sized object image which would conduct better results. Another parameter is dynamic range since the robot is working outdoors and the sensor should be able to tolerate a broad range of lighting. The robot is working in real-time so the speed of capturing images should be fast.
    
    With respect to the dynamic range parameter, there are two common global shutter CMOS sensor types for industrial cameras which are Sony Pregius\footnote{$https://www.sony\-semicon.co.jp/products\_en/IS\\/sensor0/industry/
    technology/pregius.html$} and ON semiconductor Python\footnote{$https://www.ptgrey.com/on\-semi\-python$}. The Sony Pregius sensors have a high dynamic range, whereas Python sensors are fast. In this case, having a high dynamic range is more important than the speed since the speed of detection methods is slower than the sensor speed and the current sensors have a suitable speed for real-time applications. Hence, an affordable range of cameras is selected with Sony IMX sensors for which the dynamic range is more than 65 dB. Among cameras with USB 3.0 and GigE interfaces, USB 3.0 with a global shutter is selected because it can send data securely, quickly and reliably from the camera to PC.
    
\begin{figure}[!htbp]
        \centering
            \includegraphics[width=3.3in, height=6.7cm]{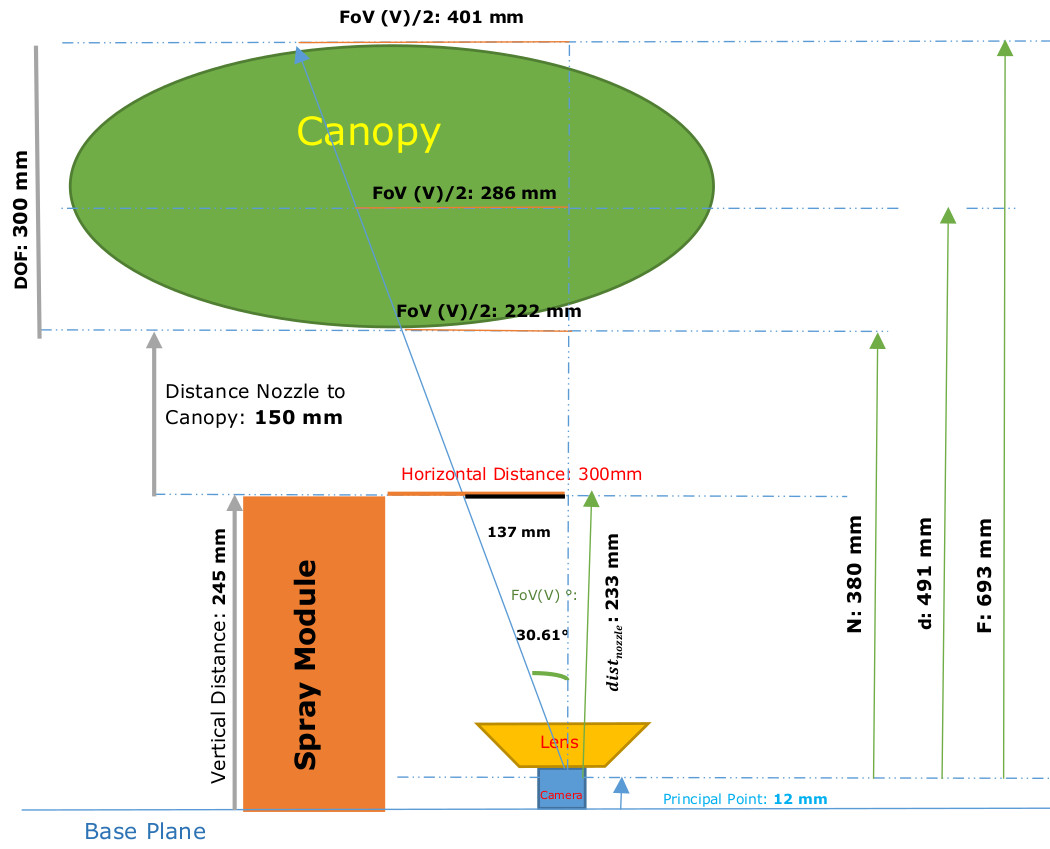}
        \caption{ The geometry of the camera related to the spray module using acA1920-40uc camera and LM6HC lens}
        \label{label3}
    \end{figure}

    To be sure that a suitable camera and lens are selected, the camera should capture sharp images at the desired distance and should be fit in the camera placement space. A view of the placement of the final selected camera and the spray module is shown in figure \ref{label3}. The measurements in the figure will be explained in the rest of the paper. Our assumption is using the pinhole camera model to simplify the computation. 
    
    The maximum horizontal distance between the camera and the nozzle is 400 mm due to the design constraints of the robot. The vertical distance between the camera and the canopy is 380 mm. It means the camera should work reliably for flowers with a distance of 380mm or more.
    
    There are a list of variables that would be used for camera selection and placement. The name of these variables, their units and description is shown in table \ref{table1}.
    
    \begin{table}[!h]
        \caption{Camera selection parameters}
        \label{table1}
        \begin{center}
            \begin{tabular}{|c||c|c|}
                \hline
                Variables & Description & Unit\\
                \hline
                $FoV$ & Field of view & $mm$ \\
                \hline
                $np$ & \multicolumn{1}{c|}{\begin{tabular}[c]{@{}c@{}}Number of pixels shows\\ the stigma area \end{tabular}}& $px$\\
                \hline
                $r$ & Image resolution & $px$\\
                \hline
                $ss$ & Size of the stigma area & $mm$\\
                \hline
                $d$ & Working distance & $mm$\\
                \hline
                $s$ & Sensor size of the camera & $mm$\\
                \hline
                $f$ & Focal length of the lens & $mm$\\
                \hline
                $f_{pixel}$ &Focal length of the lens& $px$\\
                    &in pixel unit&\\
                \hline
                $FoV(H)$ & Horizontal field of view & $mm$\\
                \hline
                $FoV(H)^\circ$ & Horizontal field of view & $^\circ$\\
                \hline
                $DoF$ & Depth of field & $mm$\\
                \hline
                $F$ & DoF far limit  & $mm$\\
                \hline
                $N$ & DoF near limit & $mm$\\
                \hline
                $H$ & Hyperfocal distance & $mm$\\
                \hline
                $f_{stop}$ & F-stop of the lens & --\\
                \hline
                $CoC$ & Circle of confusion & $mm$\\
                \hline
                $FoV(V)$ & Vertical field of view & $mm$\\
                \hline
                $FoV(V)^\circ$ & Vertical field of view & $^\circ$\\
                \hline
                $dist_{nozzle}$ &\multicolumn{1}{c|}{\begin{tabular}[c]{@{}c@{}}Distance of the top point \\of the nozzle from \\the principal point \end{tabular}}&$mm$\\
                \hline
                $b$ &Baseline& $mm$\\
                \hline
                $\epsilon_{depth}$ &Depth error of matching& $mm$\\
                \hline
                $\epsilon_{disparity}$ &\multicolumn{1}{c|}{\begin{tabular}[c]{@{}c@{}}Matching error \\(disparity values) \end{tabular}}& $px$\\
                \hline
                $b_{upper}$ &Upper bound of the baseline &$mm$\\
                \hline
                $w$ &Required overlap fraction & --\\
                \hline
                $b_{lower}$ &Lower bound of & $mm$\\
                 &the baseline &\\
                \hline
                $dv_{max}$ & Maximum number of & $px$\\
                 & disparity value &\\
                \hline
                $sd$ & Sensor's diagonal size & $mm$\\
                \hline
            \end{tabular}
            \end{center}
            \begin{tablenotes}
                \item
                \footnotesize Note: $mm$ refers to millimeter, $px$ refers to pixel, $^\circ$ refers to degree 
            \end{tablenotes}
    \end{table}
    To calculate the measurements that are shown in figure \ref{label3}. First, we need to calculate the resolution of the camera ($r$) based on a simple proportion. The size of the stigma area ($ss$) 
    is to the number of pixels presents it in the image ($np$) as the resolution ($r$) is to the field of view ($FoV$):
    \begin{equation}
        \label{eq:resolution}
       \mathit{ r = FoV \times \frac {np} {ss} }
    \end{equation}
    Equation (\ref{eq:resolution}) can be applied to both horizontal and vertical FoV which gives the width and height of the image resolution. In the rest, the width of the image resolution is computed. The width of the spray module which the camera should cover is 500 mm. In other words, the horizontal FoV(H) should be 500mm. The width and height of the stigma ($ss$) in the real world are $\sim$30 mm. The current detection method can detect the stigma area with a width and height of $>$62 px at the working distance ($np$). This implies the horizontal resolution of the camera at the working distance using (\ref{eq:resolution}) should cover greater than $\sim$1166 px. 
    
\begin{figure}[!htbp]
        \centering
                \includegraphics[width=3.2in, height=3.5cm]{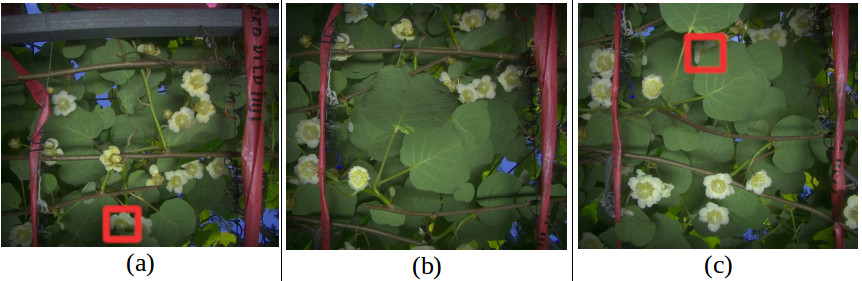}
        \caption{ The visibility of a flower when the robot moves underneath the canopy. The flower is a) partially visible b) is invisible c) partially visible and the stigma area is occluded }
        \label{fig:occlusion}
    \end{figure}

    At least the camera should capture each flower once before the robot moves out of the view. Orchard obstructions and leaves can cause flower occlusion. Figure \ref{fig:occlusion} shows an example of the visibility of a flower when the robot is moving underneath the canopy in three different frames that the camera view covers the flower location.
    
    In each frame, the camera captures the image from a different view. It can be seen in figure \ref{fig:occlusion}, the flower is invisible in \ref{fig:occlusion}-b) and is partially visible in \ref{fig:occlusion}-a). Due to the complexity of the orchard, it may be some flowers which are not visible in any three different views which is unavoidable. Based on a rule of thumb, we assume if a flower is captured from three different views, it will be partially visible from at least one view. 
    
    The maximum velocity of the robot is 1.4 m/s and we assume the processing time of the flower detection and localisation method is 100 ms per image. Therefore, the vertical field of view should be greater than 140 mm. Due to our assumption that the flower should be capture in three frames, FoV(V) is 420 mm. By using (\ref{eq:resolution}), the height of resolution should be greater than 882 px.
    
    A list of cameras is selected to easily explain the camera selection procedure. There are several brands for cameras, but usually, cameras with the same features can be found in each brand. We selected some Basler cameras with USB 3.0 interface and Sony IMX sensor which have different resolution and sensor size greater than $1166\times882$ px as is shown in table \ref{table2}. 
    
    \begin{table}[h]
        \caption{A list of candidate cameras with different image resolution and sensor size}
        \label{table2}
        \begin{center}
            \begin{tabular}{|c||c|c|c|}
                \hline
                \# & Camera type& Resolution & Sensor size\\
                \hline
                1 & acA1440-220uc & 1440$\times$1080 & 5$\times$3.7\\
                \hline
                2 & acA1920-40uc & 1920$\times$1200 & 11.3$\times$7.1 \\
                \hline
                3 & acA2040-55uc & 2048$\times$1536 & 7.07$\times$5.30  \\
                \hline
                4 & acA2440-35uc & 2448$\times$2048 & 8.4$\times$7.1\\
                \hline
                5 & acA3080-57uc & 3088$\times$2064 & 7.4$\times$5 \\
                \hline
            \end{tabular}
        \end{center}
    \end{table}
    
    To select the best camera, we need to define some constraints and choose the best option which can meet all the requirements. 
    
    It is required to have a sharp image in the depth of field (DoF). It can be seen from figure \ref{label3}, the camera should be mounted on the same base plane as the spray module. According to the nature of the kiwifruit orchard and the robot design, the height of the spray module from the base plane is 245 mm, and the distance from the top of the spray module to the canopy is defined as 150 mm. In addition, DoF of the kiwifruit flower canopy is 300 mm (figure \ref{label3}). As a result, the desired focus range of the robot is [395,695], and ideally, the working distance from the base plane is 508 mm. Since the dimension of all cameras is the same (29.3$\times$29$\times$29) mm, the principal point of the camera is 12 mm upper than the base plane (figure \ref{label3}). Thus, the ideal range of distance to the objects changes to [383,683].
    
    Another parameter that is helpful to select a suitable camera is the working distance. The working distance ($d$) for the pinhole camera model is calculated by:
    
    \begin{equation}
        \label{eq:workingDistance}
        \mathit{d =  \frac {FoV(H) \times f} {s} }
    \end{equation}
    
    To calculate $d$ for candidate cameras, the focal length of the lens ($f$) is required besides camera properties. Two lenses were selected with 4 mm and 6 mm focal length to simplify the selection. Lenses with smaller focal length have a high distortion which is not acceptable.
    
    The sensor size and resolution of cameras were extracted from the data-sheet and shown in table \ref{table2}. For each camera from table \ref{table2}, $FoV(H)$ and $d$ with presuming two different focal lengths are computed using (\ref{eq:resolution}) and (\ref{eq:workingDistance}), respectively and are shown in table \ref{table3}.

    \begin{table}[h]
        \caption{Horizontal field of view ($FoV(H)$) and working distance ($d$) for candidate cameras and lenses. $d(x)$: is the working distance while a lens with $x$ $mm$ focal length is used, each number in the first column corresponded to the cameras in table \ref{table2}}
        \label{table3}
        \begin{center}
            \begin{tabular}{|c||c|c|c|}
                \hline
                \#& $FoV(H)$ & $d$(4) &$d$(6)\\
                \hline
                1  & 696.7 &557.4 &836.1\\
                \hline
                2& 929.0 & 328.8 &493.2\\
                \hline
                3& 990.9 & 560.6 &840.9\\
                \hline
                4 & 1184.5 & 560.7 &841.0\\
                \hline
                5& 1494.1 & 807.6 & 1211\\
                \hline
            \end{tabular}
        \end{center}
    \end{table}
    
    As discussed earlier, the ideal working distance for this application is approximately 508 mm. The reason for not selecting the lens with focal length more than 6 mm is that increasing the focal length will increase the working distance which would exceed the limitations for all cameras except number 2 in table \ref{table2}. Number 5 camera is discarded since its d value in table \ref{table3} is out of range.
    
    As mentioned earlier, capturing a sharp image of the canopy is one of the essential factors of selecting cameras, and the desired range is [383,683].
    
    The definition of DoF is the range that the object appears sharply in the image. DoF is the difference between DoF far limit (F) and DoF near limit (N), as follows: 
    
    \begin{equation}
        \label{eq:DoF}
        \mathit{DoF =  F-N }
    \end{equation}
    
    The distance between the camera and the closest/furthest object that is acceptably sharp in the image is called the DoF far/near limit. In order to calculate DoF, $F$ and $N$ are required. These parameters are calculated via the following equations \cite{Potmesil2002}:
    
    \begin{equation}
        \label{eq:N}
        \mathit{N =  \frac {H\times d} {H+d-f}}
    \end{equation}
    \begin{equation}
        \label{eq:F}
        \mathit{F =  \frac {H\times d} {H-d+f}}
    \end{equation}
    
    Hyperfocal distance ($H$) is the closest distance that the lens can be focused while the object is at infinity and the image is acceptably sharp which is computed as follows:
    
    \begin{equation}
        \label{eq:H}
        \mathit{H = \frac {f^2} {f_{stop} \times CoC}}
    \end{equation}
    
    The f-stop of the lens ($f_{stop}$) is the ratio of the system’s focal length to the diameter of the aperture. A lens with a smaller f-stop projects brighter images and requires a smaller exposure time. In theory, the detection method is able to achieve higher performance in bright images. The two smallest standard f-stop for the given lens are 1.8 and 2.8. In reality, even the best lenses are not able to focus perfectly. This means the lens produces a spot instead of a point and the smallest spot is called the circle of confusion ($CoC$). Assuming the diagonal measure of the camera sensor ($sd$) is given, the circle of confusion is computed using Zeiss formula ($\frac{sd}{1730}$)\cite{Young2015}. On the basis of this criterion, $H$ is computed for all eligible cameras of table \ref{table3} with 1.8 and 2.8 f-stops. The calculated $H$ is used to calculate $F$ and $N$ (using (\ref{eq:N} and \ref{eq:F}) which are shown in table \ref{table4}. All computed parameters in table \ref{table4} used 2.8 f-stop due to computed $N$ and $F$ with f-stop 1.8 are out of range.
    
    \begin{table}[h]
        \caption{
       DoF near limit, working distance and far limit for candidates cameras and lenses. N(x), d(x), and F(x): are the distances while a lens with $x$ $mm$ focal length is used}
        \label{table4}
        \begin{center}
            \begin{tabular}{|c||c|c|c|c|c|c|}
                \hline
                \# & $N$(4) & $d$(4) & $F$(4) & $N$(6) &$d$(6) &$F$(6) \\
                \hline
                1 & 413.4 & 557.4 & 855.2 & 678.5 &836.1 &1088.9 \\
                \hline
                2 & 228.6 & 328.8 & 585.7 & 381.6 &493.2 &697.1 \\
                \hline
                3 & 372.9 & 558.2 & 1109.4 & 629.0 &837.4 &1252.1 \\
                \hline
                4 & 346.0  & 560.7 & 1477.3 & 594.9 &841.0 &1434.4  \\
                \hline
            \end{tabular}
        \end{center}
    \end{table}
    
    As mentioned earlier, the desired working range is [383,683]. From the listed camera in table \ref{table4}; $N$ and $F$ of number 1 are out of the requested range. Cameras number 3 and 4 with a 4 mm focal length and camera number 2 with a 6 mm focal length are suitable. Cameras with 4 mm focal length have a higher distortion compared to the lens with a 6 mm focal length. Therefore, the best combination of camera and lens is the Basler acA1920-40uc\footnote{https://www.baslerweb.com/en/products/cameras/area-scan-cameras/\\ace/aca1920-40uc/} camera with a 6 mm focal lens. LM6HC\footnote{https://lenses.kowa-usa.com/hc-series/417-lm6hc.html} lens is chosen which has F1.8, 1" format and -0.2\% distortion. In order to be confident that the combination satisfies requirements, we need to check other parameters as well.
    
    The next step is finding a suitable horizontal distance between the camera and the spray module. The spray module should not be visible in the image. The visible vertical FoV ($FoV(V)$) is calculated based on the tangent function by:
    
    \begin{equation}
        \label{eq:FOV(V)}
        \mathit{FoV(V) = tan(\frac {FoV(V)^\circ} {2}) \times dist_{nozzle}}
    \end{equation}
    
   According to the lens data-sheet, the ($FoV(V)^\circ$) is 61.22$^\circ$. Distance between the top of the nozzle and the principal point of the camera ($dist_{nozzle}$) is 233 mm (figure \ref{label3}). On the basis of this criterion, $FoV(V)$ of the camera at $dist_{nozzle}$ would be 137.8 mm. It should not be further than 400 mm from the spray module according to the robot design. Thereby, the range of the horizontal distance between the camera and the spray module should be in the range of [137,400]. A floodlight bar from PLR20 series with 40 LEDs, white light and 200kw was chosen to moderate the uneven lighting. The light bar is able to distribute equally the white light in the working distance. In the end, the horizontal distance is considered as 300 mm due to the required space for the LED light bar. All calculated parameters for the selected camera and lens are shown in figure \ref{label3}.
    
    Other parameters that are worthwhile to mention are the vertical and horizontal FoV and the number of pixels that show the stigma ($np$) at a different distance as shown in table \ref{table5}.
    
    \begin{table}[h]
        \caption{FoV (V) and FoV(H) and the number of pixels presents the stigma ($np$) in near/far focus limit and working distance ($d$) using (\ref{eq:resolution}) and (\ref{eq:workingDistance})}
        \label{table5}
        \begin{center}
            \begin{tabular}{|c||c|c|c|c|c|c|}
                \hline
                Parameter & $N$ & $d$ & $F$  \\
                \hline
                $FoV(V) (mm)$ & 450.09 & 581.41 & 820.91 \\
                \hline
                $FoV(H) (mm)$ & 718.25 &  927.8 & 1310\\
                \hline
                $np$ & 80 & 62 & 44 \\
                \hline
            \end{tabular}
        \end{center}
    \end{table}
    
    The $FoV(V)$ at working distance is 581 mm and based on our assumption a flower should be seen at least in 3 images. Consequently, if the robot moves in the highest speed (1.4 m/s), the camera should cover 193mm ($\frac{581}{3}$) and the software modules should process an image in less than 137ms. The stigma area at the working distance is shown by 62 px, and this number drops to 44 px at the far distance. Therefore, it can be expected that some flowers at the far distance may not be detected.

   \subsection{Finding the Baseline} 
\begin{figure}[!htbp]
        \centering
            \includegraphics[width=3in, height=5cm]{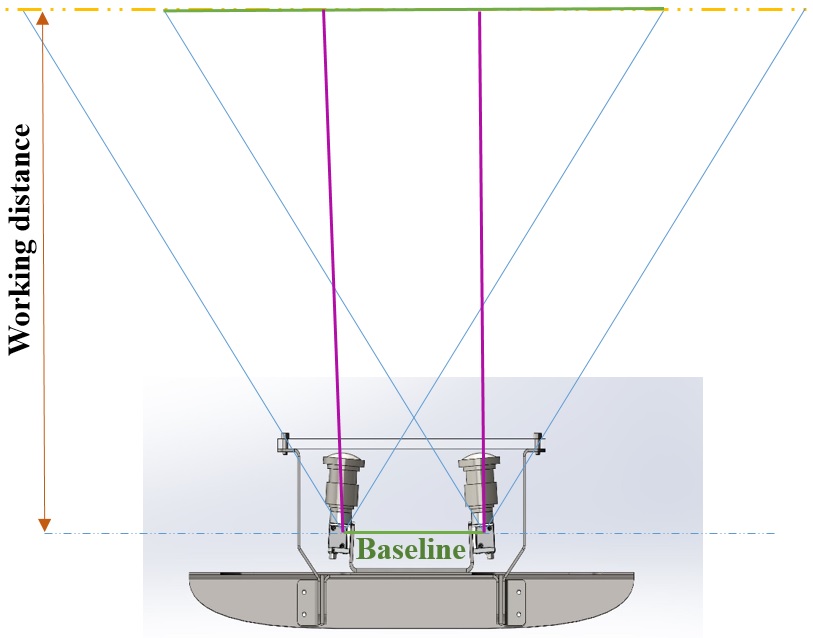}
        \caption{Structure of the stereo cameras}
        \label{label4}
    \end{figure}

    After selecting the proper camera, its baseline should be calculated. The baseline is the distance between the two cameras (figure \ref{label4}). The centre of the baseline is aligned with the centre of the nozzle.
    
    The baseline length leads to a trade-off, and a large baseline provides a small matching error. Conversely, a small baseline gives a large matching error. Some constraints are defined for finding the best baseline. The depth error of stereo matching ($\epsilon_{depth}$) can be calculated by \cite{Gallup2008}:
    
    \begin{equation}
        \label{eq:depth}
        \mathit{\epsilon_{depth} = \frac {z^2} {b \times f_{pixel}} \times \epsilon_{disparity}}
    \end{equation}
    
    Where $\epsilon_{disparity}$ is the matching error and relevant to the matching method. In our assumption, $\epsilon_{disparity}$ is 1 pixel. The depth error increases quadratically with the distance of the object from the camera ($z$). Moreover, depth error ($\epsilon_{depth}$) and baseline ($b$) are inversely related to each other. The constraint on the spray module indicates that $\epsilon_{depth}$ must be less than 3 mm. On the basis of this criterion, using (\ref{eq:depth}) the lower bound at the working distance for the baseline is 149 mm.
    
    Besides, the overlap between the two cameras is related to the baseline \cite{Gallup2008}. The upper bound of the baseline $b_{upper}$ is calculated by:
    
    \begin{equation}
       \mathit{ b_{upper} = 2 \times f \times tan(\frac {FoV(H)^\circ}{2}) \times [1-w]}
    \end{equation}
    
   Where $w$ is the required overlap fraction. At the working distance, the desired $FoV(H)$ is 500 mm (the spray module covers 500mm) and $FoV(H)$ of one camera is 927 mm (\ref{table5}), and $w$ is calculated as $\frac {500}{927}=0.53$. As a result, the upper bound of the baseline is 427 mm, given the horizontal field of view at the working distance ($FoV(H)$) and the required overlap fraction ($w$). Consequently, the baseline should be in the range of [149,427].
   
   On the other hand, a large baseline increases the maximum disparity and occlusion. The disparity affects the size of the search window in the stereo matching method, which is dependent on the processing time. The limitation on the maximum disparity is 500 pixel. The lower baseline $b_{lower}$ can be calculated using the maximum disparity:
   
    \begin{equation}
       \mathit{ b_{upper} = \frac {d \times dv_{max}}{f}}
    \end{equation}
    
\begin{figure}[!htbp]
        \centering
                {\includegraphics[width=3in, height=5cm]{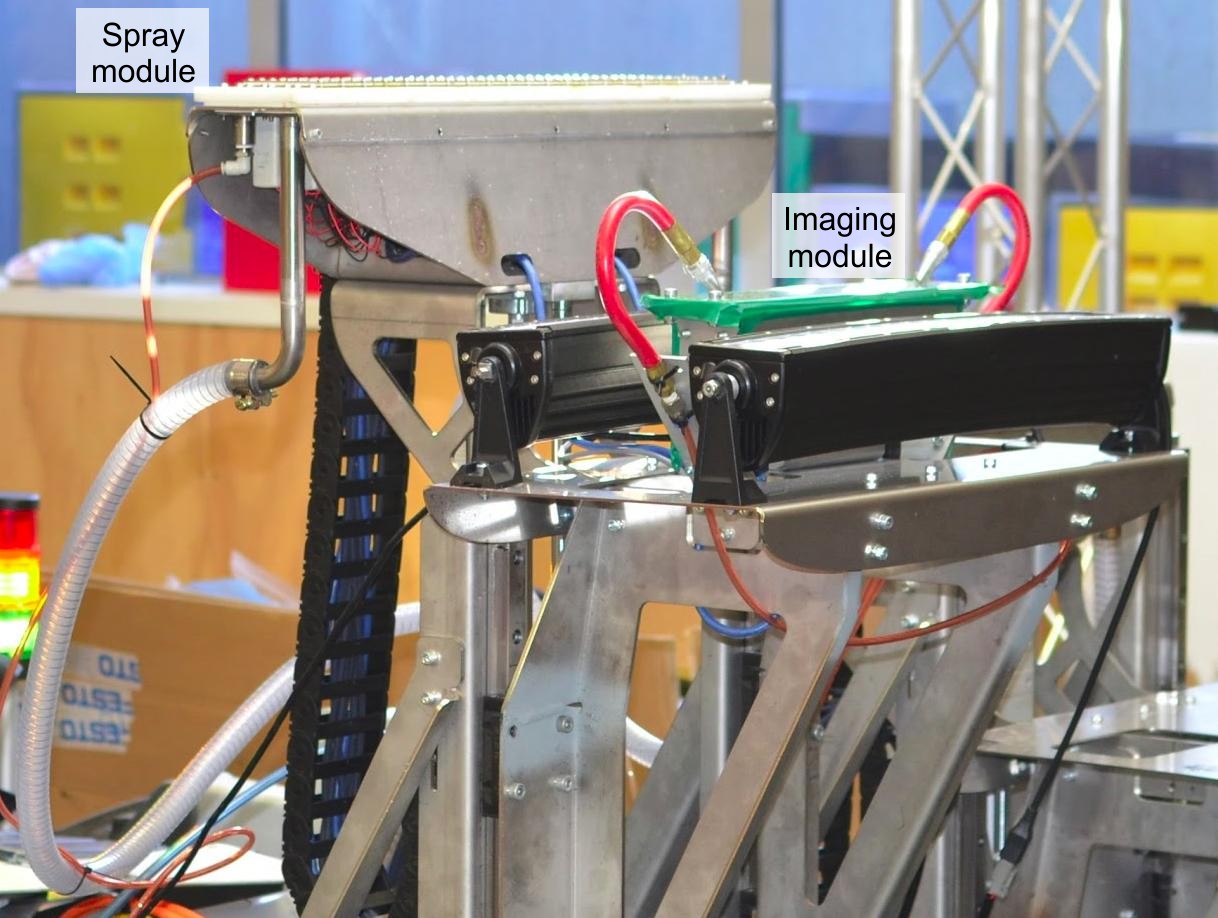}}
        \caption{The kiwifruit flower pollinator platform}
        \label{label5}
    \end{figure}

    Where $d$, $dv_{max}$, and f are 508mm, 500 pixel, and 6mm, respectively. With the current setup, the baseline should be in the range of [149,187], so we choose the mean number which is 168 mm. However, in the building process of the module, this number changed to 170 mm which is still acceptable. The hardware part of the imaging module with the discussed measurements was built and integrated to the spray module is shown in figure \ref{label5}.
    
    \section{Experiments}\label{sec:Experiments}
    As we discussed earlier, stereo cameras are suitable for real-time outdoor scenes compared to other sensors; However, the disadvantages are capturing overexposed and underexposed images and requiring camera calibration. The performance of the sensing module was measured and evaluated in the real-world based on the visibility, quality of taken images, calibration error, and the robot performance in the orchard.
    
    \subsection{Images with Unpredictable Lighting Conditions}
\begin{figure}[!htbp]
        \centering
        \includegraphics[width=3.2in, height=6cm]{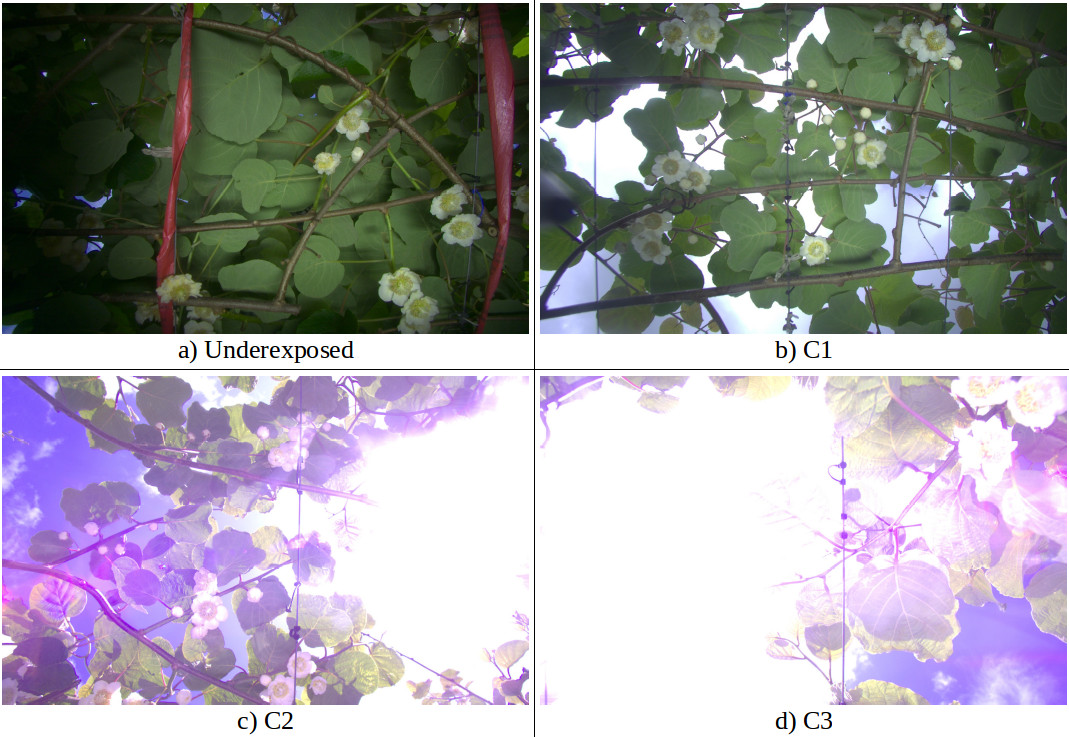}
        \caption{An example of a) an underexposed image, b) typical image (category C1), c,d) glare images (categories C2 and C3)}
        \label{figOverund}
    \end{figure}

    One of the disadvantages of using colour cameras is having underexposed, and glare images. An example of these images is shown in figure \ref{figOverund}. 
    Glare happens when there is a layer of purplish colour on all pixels. 
    
    Regarding the underexposed image, the width of the light bar was the same as spray module (500mm) so it caused darkness in the left side of the left image and the right side of the right image. These parts were not contributed in stereo matching so it can be disregarded. 
    With regard to glare images, the probability of capturing them depended on the foliage coverage, the sun angle, and the weather. It happened when the camera faces the sun and there was a low foliage coverage. Thus, the sun caused some flowers became partially visible.
    
    Consequently, glare images would be captured when it was sunny weather and early season. Visually, a layer of purple colour might be seen on the majority of the glare images. It seemed a high number of pixels should be saturated in the RGB channel and especially B channel. A parameter is defined to measure the probability of having this kind of images which is called saturation rate. The rate of saturated pixels over the image size was counted for each image and it is called the saturation rate (SR) and the range is [0,100]. Three classes were defined to categorise the status of images according to the percentage of saturated pixels. Category C1 indicates the typical images which less than 25\% of pixels are saturated (figure \ref{figOverund}-b). The flowers in images of class C1 are totally visible. Category C2 presents the images which the percentage of saturated pixels in the image are between 25\% to 50\% (figure \ref{figOverund}-c). Hence, some regions would be affected by the sun and some flowers might be partially visible. Category C3 refers to the images which more than 50\% of pixels are saturated and some flowers might be invisible (figure \ref{figOverund}-d).
    
    As can be seen in table \ref{tableOverImg}, a dataset from four different orchards was captured during the daytime with sunny and forecast weathers in different season time. The Bateman dataset was captured in the early season and included some closed flowers and less foliage coverage.
    
    In the table \ref{tableOverImg}, the images were captured from Haukiwi, Steve and Newnham orchards in sunny and cloudy weather were almost typical and the variation lighting did not affect them. However, around 20\% of images of Bateman orchard belonged to category C2 and about 2\% are from category C3. The reason is the data captured in early season from Bateman orchard which the foliage coverage was sparse and even some flowers were closed. In addition, in our assumption, each non-occluded flower was captured in 3 consecutive frames. In Bateman set, only 8\% of images were from category C2 or C3 which its 3 consecutive frames were also from category C2 or C3. This indicates there is a low chance that a flower can be invisible due to the unpredictable lighting conditions.
    
    \begin{table}[h]
        \caption{The percentage of captured images from Haukiwi, Steve, Newnham, and Bateman orchards belongs to C1, C2, and C3, Img refers to the number of images}
        \label{tableOverImg}
        \begin{center}
            \begin{tabular}{|c||c|c|c|c|c|}
                \hline
                Orchard's & Weather&Img & $C1$ & $C2$  &
                $C3$ \\
                name&&&(\%)&(\%)&(\%)\\
                \hline
                Haukiwi & Sunny & 420&100 & 0.0 & 0.0  \\
                \hline
                Steve & Cloudy & 5622&99.6 &  0.4 &  0.0  \\
                \hline
                Newnham & Sunny & 1610&100 & 0.0 & 0.0 \\
                \hline
                Bateman & Sunny &  1476&78.0 & 20.1 & 1.9 \\
                \hline
                Overall & - &  9128 & 96.1 & 3.5 & 0.4 \\
                \hline
            \end{tabular}
        \end{center}
    \end{table}
    
    
    \subsection{Camera Calibration}
    The camera should be calibrated to obtain 3D positions. The standard Opencv stereo calibration library \cite{Zelinsky2008} with pinhole camera model was used to calibrate cameras which were hardware-triggered. 70 Images of a checkerboard with 6$\times$8 and 60mm square size were captured. The images were covered the whole field of view in different distance from the camera. The tangential and radial distortions were considered in the calibration. The output of the calibration is the extrinsic and intrinsic matrix which include the internal and external camera parameters.
    
    The accuracy was measured based on the RMS error and epipolar error, which were 0.37 px and 0.19 px, respectively. The RMS error is the distance between the corners of checkerboard and the projected points and as a rule-thumb, RMS error should be less than 1 px. The epipolar error is the distance between the corner's epipolar line on the left image to its correspondence point in the right image. The RMS and epipolar errors in the calibration were 0.35 px and 0.19 px, respectively. Moreover, the error in x and y direction were calculated to have a more realistic estimation about the error. The x-direction and y-direction errors were less than 3mm. Accordingly, they can meet the criteria for sensing modules. All errors are shown in table \ref{table6} and it seems all errors are reasonable for the application.
    
    \begin{table}[h]
        \caption{The calibration error}
        \label{table6}
        \begin{center}
            \begin{tabular}{|c||c|c|c|c|c|c|}
                \hline
                RMS & Epipolar & X-direction & Y-direction\\
                error(px)& error(px)&error (mm)&error (mm)\\
                \hline
                \hline
                0.37 & 0.19 & 2.98 & 1.28681 \\
                \hline
            \end{tabular}
        \end{center}
    \end{table}
    
    \subsection{The Robot Performance In the Orchard}
    The kiwifruit flower pollinator platform was mounted on the robot and tested in the real orchard. The constraints on testing in the real environment were the short blooming season, unpredictable weather and limitation of orchard access. The spray module was not tested in nighttime due to the wet pollen was used for pollination. With using wet pollen, the germination process works efficiently in warm conditions \cite{Jansson1988}. In view of the fact that the daytime is warm, the robot was tested during the daytime.
    
    The robot was tested in the kiwifruit flower orchard and the number of pollinated flowers were counted manually. A red dye was used with the wet pollen to make visible the pollinated flowers. The counted flowers were placed between two wires with 400-500 mm wide as shown in figure \ref{label2}. The hit rate was the number of hit flowers out of all flowers. A low number of flowers were invisible from the camera view because of the obstruction. The unreachable flowers are the visible ones by the camera but the obstruction on the way of shot pollen avoided the pollen to reach the stigma.
    
    The velocity, hit-rate, the number of flowers, and the number of images that a flower can be seen in normal conditions from different angles were considered for evaluation which is shown in table \ref{table7}. In our assumption in normal conditions a flower is not occluded from the camera view and the speed of the imaging module is 10 fps. In the highest speed (5 km/hr), the robot moves 1388 mm per second and we assumed the whole imaging module works 10 fps. Therefore, the robot moves forward 138mm in each frame. The FoV(H) based on table \ref{table5} in near, working, and far distances were 450.9mm, 581.41mm, and 820.91mm, respectively. In the worst case, a flower in the near distance can be seen at least in 3 frames ($\frac{450.9}{134}$). The visibility in the lowest speed and higher distance is higher due to a flower can be seen from more different views. The low hit rate was caused by the low flower detection and localization performance, unreachable flowers, and compatibility between the delay of the spray and imaging module. The best result among all tested velocities was achieved at 3.5 km/h with a 79.5\% hit rate. \cite{Williams2019}. 
    
    \begin{table}[h]
        \caption{The hit rate results. Visibility indicates the number of images from a flower is captured in normal conditions}
        \label{table7}
        \begin{center}
            \begin{tabular}{|c||c|c|c|c|c|c|}
                \hline
                Velocity& Flowers & Hit rate & Visibility  \\
                (km/hr)&&(\%)&\\
                \hline
                1 & 731 & 74.0{\mypm5.8} & 16\\
                \hline
                1.5 & 931 &  70.1{\mypm11.0} & 10\\
                \hline
                2.5 & 1171 & 75.5{\mypm5.3} & 6 \\
                \hline
                3.5 & 1102 & 79.5{\mypm3.9} & 5 \\
                \hline
                5 & 1231 & 56.0{\mypm5.6} & 3 \\
                \hline
            \end{tabular}
        \end{center}
    \end{table}

    \section{Conclusion}\label{sec:conclusion}
    A sensing module was designed for a kiwifruit flower pollinator to be integrated with an imaging module and a spray module. Some constraints based on the orchard environment and the robot design were defined. The limitations are DoF, FoV, camera placement space, working distance, maximum depth error, maximum disparity, and the conditions of working in real-time and an outdoor scene. The Basler ac1920-40uc USB 3.0 camera with Kowa lenses (LM6HC) was chosen among five different cameras and lens. In order to find the correct camera placement, the limitations on FoV, DoF and working distance are considered. Two cameras were used due to the requirement to find the 3D position of the target. The baseline is defined according to the limitations on maximum disparity, depth error and FoV. 
    
    With purpose of evaluating the designed sensing module, sensitivity to changeable lighting, the calibration error and the robot performance were discussed. The sensing module was robust to dynamic lighting conditions in mid/late seasons. The only condition was the sunny weather in early season which the sun faces the cameras and it might cause some flower to become invisible which happened in only 8\% of captured images. The current calibration was good enough for the requirement. One of the suggestion to increase the accuracy of calibration would be detecting the corners of the checkerboard using a convolutional neural network which is more accurate \cite{Donne2016}. The imaging module in the highest speed can capture a flower in three different frames from different views which decrease the probability of invisibility. The imaging module was tested on a robot in a kiwifruit flower orchard, and it was able to hit 79.5\% of targets in 3.5 km/h.

\section*{Acknowledgments}
    
    This research was funded by the New Zealand Ministry for Business, Innovation and Employment (MBIE) on contract UOAX1414.

    
    
\end{document}